\documentclass[runningheads]{llncs}
\usepackage{graphicx}
\usepackage{url}
\usepackage{float}
\usepackage{array}
\usepackage{caption}
\usepackage{subcaption}
\usepackage{etoolbox}
\usepackage[table]{xcolor}
\renewcommand{\arraystretch}{2.5}
\usepackage{enumitem}
\newlist{steps}{enumerate}{1}
\setlist[steps, 1]{label = Step \arabic*:}

\usepackage{adjustbox}
\usepackage{background}
\backgroundsetup{contents={DRAFT}, opacity=0.20, scale=22}

\begin{document}
\title{SPRITZ-PS: VALIDATION OF SYNTHETIC FACE IMAGES USING A LARGE DATASET OF PRINTED DOCUMENTS
}
%
%
\author{Ehsan~Nowroozi 
\inst{1} 
, Yoosef Habibi\inst{2}, and Mauro Conti\inst{2}}
\authorrunning{E. Nowroozi et al.}

\institute{Department of Computer Engineering, Bahcesehir University, Istanbul Turkey.\\
\email{ehsan.nowroozi@eng.bau.edu.tr}\\
\and
Department of Mathematics, University of Padua, Via Trieste, 63 - Padua, Italy\\
\email{yoosef.habibi@studenti.unipd.it, mauro.conti@unipd.it} 
}
\maketitle              

\begin{abstract}
The capability of doing effective forensic analysis on printed and scanned (PS) images is essential in many applications.  PS documents may be used to conceal the artifacts of images which is due to the synthetic nature of images since these artifacts are typically present in manipulated images and the main artifacts in the synthetic images can be removed after the PS. Due to the appeal of Generative Adversarial Networks (GANs), synthetic face images generated with GANs models are difficult to differentiate from genuine human faces and may be used to create counterfeit identities. Additionally, since GANs models do not account for physiological constraints for generating human faces and their impact on human IRISes, distinguishing genuine from synthetic IRISes in the PS scenario becomes extremely difficult. As a result of the lack of large-scale reference IRIS datasets in the PS scenario, we aim at developing a novel dataset to become a standard for Multimedia Forensics (MFs) investigation which is available at {~\cite{f1fx-sq21-22}}.
\\In this paper, we provide a novel dataset made up of a large number of synthetic and natural printed IRISes taken from VIPPrint Printed and Scanned face images. We extracted irises from face images and it is possible that the model due to eyelid occlusion captured the incomplete irises. To fill the missing pixels of extracted iris, we applied techniques to discover the complex link between the iris images. To highlight the problems involved with the evaluation of the dataset's IRIS images, we conducted a large number of analyses employing Siamese Neural Networks to assess the similarities between genuine and synthetic human IRISes, such as ResNet50, Xception, VGG16, and MobileNet-v2. For instance, using the Xception network, we achieved 56.76\% similarity of IRISes for synthetic images and 92.77\% similarity of IRISes for real images.

\end{abstract}

\section {Introduction}
There are various concerns regarding the potential abuse of modern technology that are widely available for creating tangible documents, such as printers and scanners. Through these devices, various malicious activities have been conducted such as generating fake physical documents to misguide criminal investigations {~\cite{Nowroozi_survey}{\cite{Nowroozi_attacks}{\cite{Nowroozi_attacks2}\cite{NowrooziChapter2022}} . Also, to avoid monitoring, illegal documents can be printed and distributed without notice such as child pornography content. Since the artifacts presented in manipulated or even synthetic images are not visible or trackable after being printed and scanned, these devices are vastly used for harmful reasons which can also affect the economy negatively; For example, printing fake currencies and packages of products faking original ones \cite{barni2018detection}\cite{barni2017higher}\cite{nowroozi2015double}.

To primarily address this issue, color laser printer producers signed an agreement mentioning that each manufacturer should imply secret yellow dots{~\cite{1}} which are called machine identification codes (MIC). MIC is a printer steganography, unique yellow-dots patterns, applying to specify the manufacturer of the printer. But MIC can be easily modified in order not to be recognized{~\cite{Richter2018}} and also there are many printer factories which do not provide MIC. 

Considering semantic features captured from images can lead us to more robust models that their results can generalize. For detecting the fake faces that are generated by GANs, we can apply eye texture information. Recent studies revealed that some eye artifacts like the difference of colors between eyes{~\cite{Matern2019ExploitingVA}} and the inconsistencies of the corneal specular highlights{~\cite{Hu2021ExposingGF}} can be used to detect the GANs generated images. Also, there are other forensics clues that can be captured such as the iris pattern to develop more robust solutions.

Despite the research that has been done so far, the absence of substantial reference datasets is impeding advancements in this field. Except from the dataset introduced in{~\cite{VIPP}} which is also has been used for this study to provide our dataset, there are other few datasets which have at least one of the following problems:
\textbf{1.} They only include ad hoc data created for research studies.
\textbf{2.}	They include basic sample patterns such as icons, text, and halftone patterns
\textbf{3.}	Mainly provided printed documents captured from old and non-professional printers
\textbf{4.}	They lack copies of documents captured from the same printer model
\textbf{5.}	They lack complex fake printed images
Printing and scanning remove some of the most important artifacts that can be used for manipulation detection, such as the discrete cosine transform irregularities, RGB channels correlations, and also illumination inconsistencies. The lack of complex fake printed images, which is also mentioned as a flaw for mostly all available datasets. Complex fake printed images are necessary to be able to develop models that can capture remaining artifacts. Large reference datasets that address the aforementioned issues may be of great assistance in advancing new developments in printed document investigations, particularly in terms of the identification of manipulated and synthetic documents as well as for source linking specifying the printer generated the document.

\section{Related Works}
\label{Related Works}
In this section, we review the approaches in the literature related to this study. Firstly, we review GAN face synthesis and detection algorithms. Then, we discuss briefly state of the art iris detection approaches in MFs.

\subsection{GANs Application on Face Synthesis}
\label{GANs Application}

Many GAN models have been introduced in the literature in recent years for the generation and modification of face images. Face editing models, such as IcGAN{~\cite{perarnau2016}} and attGAN{~\cite{He2019}}, alter the characteristics of a face, such as the color of the skin or hair, gender, and age. In addition to face editing, another study explored the application of GAN models for face-to-face translation e.g.,{~\cite{Choi2018}} developed a model to achieve high quality translated face images. Also, there are studies focusing on generating face images from scratch i.e., synthesizing face images from random noise.

Early studies on generating face images mostly were capable of proposing models that generate high-quality, low-resolution images {~\cite{Radford2016}}{\cite{Zhao2017}}{\cite{Berthelot2017}}, where recent studies propose more realistic face images that are high-quality, large-resolution images, even up to 1,024 × 1,024.  ProGAN{~\cite{karras18}} introduced the first model capable of synthesizing face images up to (1,024 × 1,024). StyleGAN{~\cite{karras19}} applying style transfer literature to the generator architecture achieved a higher resolution image. StyleGAN considered in the generative process the ‘style’ feature as well to generate higher resolution images.  by redesigning the normalization in the generator, the StyleGAN2 model{~\cite{karras20}} has been proposed which enhanced the quality of images more than StyleGAN.

Then, StyleGAN3{~\cite{NEURIPS2021_076ccd93}} was introduced by NVIDIA. The architecture applied in the new version is totally different from StyleGAN2 and closer to the original StyleGAN architecture. As shown in{~\cite{NEURIPS2021_076ccd93}}, StyleGAN2 and StyleGAN3 learn different representations from images. StyleGAN3 enhances the quality of synthetic images which make them also suitable for being used in video and animation creation. StyleGAN3 address the issue of “texture sticking” available in the images created with StyleGAN2 by preventing of leaking the unnecessary information to be leaked into the hierarchical synthesis process.

\subsection{GANs Application on Face Detection}

To distinguish between real images and fake face images generated by GANs, various approaches are proposed in the literature. early GAN-based methods mostly applied hand-crafted facial features such as irises color or borders of face {~\cite{Matern2019ExploitingVA}}. Studies in{~\cite{Yang2019}} illustrated that extracting the locations of facial landmark points and feeding them to train an SVM is useful to determine whether an image is fake or not. Studies conducted in{~\cite{McCloskey2019}}{\cite{Li2020}} applied color information exracted from images such as the correlation between color channels, and saturation information to determine the originality of the images. Another study in{~\cite{Li2020}}, merged color channel information, specifically co-occurrences from color channels, with Subtractive Pixel Adjacency Matrix (SPAM){~\cite{5437325}}-like properties and feed them to an SVM for the detection. The authors in{~\cite{Nataraj2019}} instead of feeding color channel information to an SVM, they took advantage of Convolutional Neural Networks (CNNs) and reached a higher performance on detecting real and fake face images. In{~\cite{Barni2020}} also cross-band co-occurrence information have been fed to the CNN.

Moreover to{~\cite{Nataraj2019}} and{~\cite{Barni2020}}, many other studies{~\cite{Marra2018}}{\cite{Marra2019}}{\cite{Hsu2020}}{\cite{Hulzebosch2020}} have applied CNNs, mostly fully supervised approaches, for the fake detection. These proposed models achieve higher accuracy scores than using traditional machine learning approaches fed with the hand-crafted features. A study conducted by{~\cite{Marra2018}} showed that applying Xception, Inception and DenseNet, networks pre-trained on ImageNet and feeding the pixel images can boost the accuracy of detection.

Fully supervised models perform notably when training and test images come from same GAN models, but they lack generalization, and their performance fails when the test images are not coming from the same models that the training images are from.  There are studies to address this issue by introducing techniques for having more robust solutions e.g.,{~\cite{Liu2020}} proposed a CNN-based detector which is called Gram-Net. This model by applying global image texture representations improved the model robustness. In another study{~\cite{Chen2021}}, the authors improved the generalization by detecting the real and fake image by taking into consideration the information given from multiple semantic segments. 
Other techniques have been applied such as data augmentation to make the model learn more general representations. As an example, this study{~\cite{Xuan2019}} added augmented data performing Gaussian blurring. A similar study done by{~\cite{9156876}} provided augmentation data by applying compression and blurring and trained a pre-trained model like ResNet50. According to experiments, applying a single GAN model results in generalized solution regarding to unseen architectures, datasets, and training methods. Augmentation along with training strategies such considering residual extraction and the eliminating of down-sampling layers in first layers can increase the robustness of the model as done in{~\cite{9428429}}.

\subsection{Iris Detection in Multimedia Forensics}

The requirement for an effective human recognized proof framework in view of biometrics has increased due to the growth of apps demanding security in daily life. Moreover to the card- and password-based authentication methods, there is still a need for enhancing validation/approval systems applying new methods{~\cite{SINGH20201868}}. From cellphones and computers to border control and other sensitive sectors, biometric technologies are used pervasively. Understanding the condition of biometrics systems and how they evolved through time is crucial as many people, specialists and non- specialists are frequently in contact with these systems{~\cite{2208.09500v1}}.

Inspired by the technological advance, researchers have mainly concentrated on enhancing the deep learning-based biometric systems during the past decades. According to the results of NIST's Facial Identification Vendor Test{~\cite{Grother2018OngoingFR}}, the integration of CNNs to extract face characteristics from a face image has led to enormous improvements in face recognition performance.

Biometric frameworks{~\cite{8241848}}consist of physiological and behavioral devices. Physiological devices can be face, hand geometry, fingerprint, and iris, where behavioral clues includes  typing pattern, gait, and voice. Iris has features that distinct it from other biometrics. It provides uniqueness, stability, collectability and non-invasive. Also, Iris provides high level of texture information and more importantly, its structure has been finalized after embryonic development and cannot be modified. Among non-invasive features such as face image and voice, iris-based recognitions systems are more robust. In other words, applying iris-based recognition systems lead to lowest rate of the false recognition in comparison with other biometrics{~\cite{article}}{\cite{8237673}}{\cite{9036930}}.

For detecting fake human face images generated from GANs also iris features can be helpful to distinguish the real and fake but realistic images. John Daugman created a patented method that is used by the majority of commercial iris recognition systems{~\cite{1262028}}{\cite{491729}}. In his approach, Daugman employed the integro differential operator to calculate the iris boundaries. Yang Hu et al.{~\cite{7569033}} introduced a method for optimally generating the iris codes required for iris recognition. By using this technique, it is shown that the conventional iris code is the result of an optimization problem in which the minimum distance between the feature values and the iris codes must be achieved.

\section{SPRITZ-PS Datset}
\label{Datset}
The proposed dataset examines the identification of synthetic false images, such as those produced by GANs after being printed and scanned, which is a significant but largely unexplored issue in digital image forensics \cite{nowroozi2020machine}\cite{barni2020improving}. The proposed dataset can encourage investigations in MFs to prevent the harmful effects of utilizing fake PS images.  
To Create this dataset, we utilized the VIPPrint dataset{~\cite{VIPP}}{\cite{VIPP-dataset}} that includes human face images, which will be discussed in the next section. Then, we extracted irises from face images. The process is described in Section\mbox{~\ref{sec:Iris Extraction}}. Extracted irises might be occluded or incomplete, so we do a reconstruction phase and create our own dataset consisting of reconstructed irises from pristine and fake face images for further analysis. The procedure is described in Section\mbox{~\ref{sec:Iris Reconstruction}}. Finally, we apply an SNN to verify our novel dataset. The dataset is publicly available in{~\cite{VIPP-dataset}} and hope it encourages researchers to exploit more research using this dataset.

\subsection{VIPPrint Dataset}

VIPPrint dataset is the backbone dataset used for creating our proposed dataset.  VIPPrint dataset{~\cite{VIPP}}{\cite{VIPP-dataset}} introduced a dataset which resolve two issues of prior-existing datasets, (i) lack of diversity, and (ii) lack of redundancy. To handle the lack of diversity issue, they provided a dataset which consists of printed documents given from different models of printers with various printing resolutions. Also, they gathered documents from two or more printers of the same model and brand to address the lack of redundancy.

This dataset consists of face images which is being applied in many applications of biometric recognition and criminal investigations. Also, the fact that there are many large face datasets available that can be beneficial for further studies. besides, face is an important human attribute as progresses in AI resulted in advanced synthetic human face image generator models. This phenomenon makes it more important to have countermeasures to avoid malicious activities in MFs{~\cite{karras19}}. The face images for this dataset are originally from Flickr-Faces-HQ (FFHQ) dataset{~\cite{karras19}} which provides a large human face images. Some sample images of FFHQ dataset are shown in Figure\mbox{~\ref{fig:FFHQ}}.

\begin{figure}[hbt!]
    \centering
    \caption{Some samples given from the FFHQ dataset
}
    \includegraphics[width=0.6\textwidth]{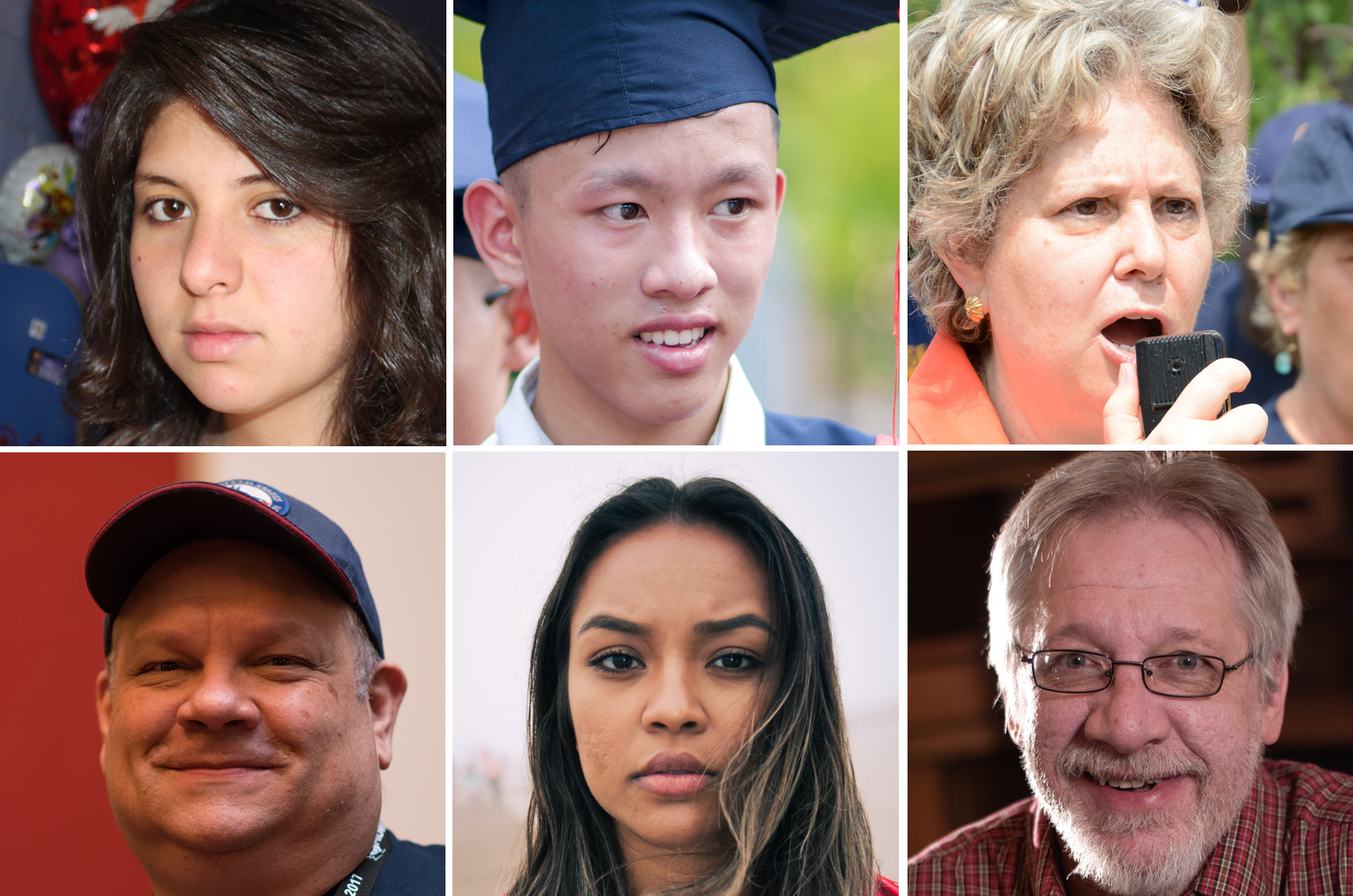}
    
    \captionsetup{justification=centering}
    \label{fig:FFHQ}
\end{figure}

They provided images given from multiple printers as reported in the Table\mbox{~\ref{tab:VIPPrint-Dataset-Printer}}. All printers are laser type, and 200 images are used from each. They printed 200 images on fifty A4 papers, each paper including 4 images, using landscape orientation. They assumed that 200 images may not be enough to deploy them for many deep learning techniques, so they provided another set of images which included Regions of Interest (ROI) extracted from the original images. The printer resolutions used for creation of VIPPrint were generally 600 × 600 dpi, while the first and last are taken differently, respectively 4800 × 2400 dpi, and 1200 × 600 dpi. The printers used for collecting printed documents were from weeks to years old. Furthermore, they scanned the documents with default configurations for sharpness, and then stored applying a lossless compression setup. The scanner used for this process was from the Kyocera TaskAlfa3551ci multifunctional printer. Comparing the HH discrete wavelet transform subbands (luminance component) differences between printers reveals that there is no big difference between images taken from same brand and model, while it is more visible comparing images taken from different printer manufacturer.

\begin{table}[hbt!]
\centering

\caption{List of printers used for creation of the first version of the VIPPrint dataset}
\label{tab:VIPPrint-Dataset-Printer}
\begin{adjustbox}{width=1.0\textwidth,center=\textwidth,height=0.12\textheight}
\begin{tabular}{|c|c|}
\hline
\textbf{Brand} & \textbf{Model}                 \\ \hline
Epson          & WorkForce WF-7715              \\ \hline
Kyocera        & Color Laser                    \\ \hline
Kyocera        & TaskAlfa 3551                  \\ \hline
Kyocera        & TaskAlfa 3551                  \\ \hline
Samsung        & Multiexpress X3280NR           \\ \hline
HP             & Color LaserJet Pro rfp-r479fdw \\ \hline
HP             & Color LaserJet rfp-r377dw      \\ \hline
OKI            & C612 LaserColor                \\ \hline
\end{tabular}
\end{adjustbox}
\end{table}

VIPPrint dataset also includes huge number of natural and GAN-generated face photos to encourage more study on this subject. Specifically, 40,000 face photos were produced and scanned using the same scanner as mentioned above with the following settings:
\begin{itemize}
\item Images created by StyleGAN2 that are 16,000 real and 16,000 synthetic{~\cite{karras19}}.
\item ProgressiveGAN produced 3500 real images and 3500 synthetic images{~\cite{karras18}}
\item 500 real and 500 fake images produced by StarGAN{~\cite{Choi2018}}
\end{itemize}

The first issue with these images is the significant pixel distortion that is produced after the printing and scanning process. Figure\mbox{~\ref{fig:ProGAN}} presents the raw pristine PS images\mbox{~\ref{fig:ProGAN-a}} and ProGAN PS-generated samples\mbox{~\ref{fig:ProGAN-b}}. Figure\mbox{~\ref{fig:StyleGAN2}} presents the raw real PS images\mbox{~\ref{fig:StyleGAN2-a}} and StyleGAN2 PS-generated samples\mbox{~\ref{fig:StyleGAN2-b}}. Also, Figure\mbox{~\ref{fig:Zoomed_StyleGAN}} is representing the zoomed regions of a real PS sample\mbox{~\ref{fig:Zoomed_StyleGAN-a}} versus a sythetic StyleGAN2 PS-generated one\mbox{~\ref{fig:Zoomed_StyleGAN-b}}. Such images having a computed Structural Similarity Index{~\cite{1284395}} of 0.41 and with a peak noise to signal ratio of 17.65 dB show that there is a severe visual degradation. As shown in the images and values from them, it is quite obvious that discriminating between printed original and GAN photos is very challenging.


\begin{figure}[hbt!]
\centering
\caption{Some raw samples and ProGAN generated samples from VIPPrint dataset} 
\begin{subfigure}{.6\textwidth}
    \caption{Raw PS samples}
    \includegraphics[width=\textwidth]{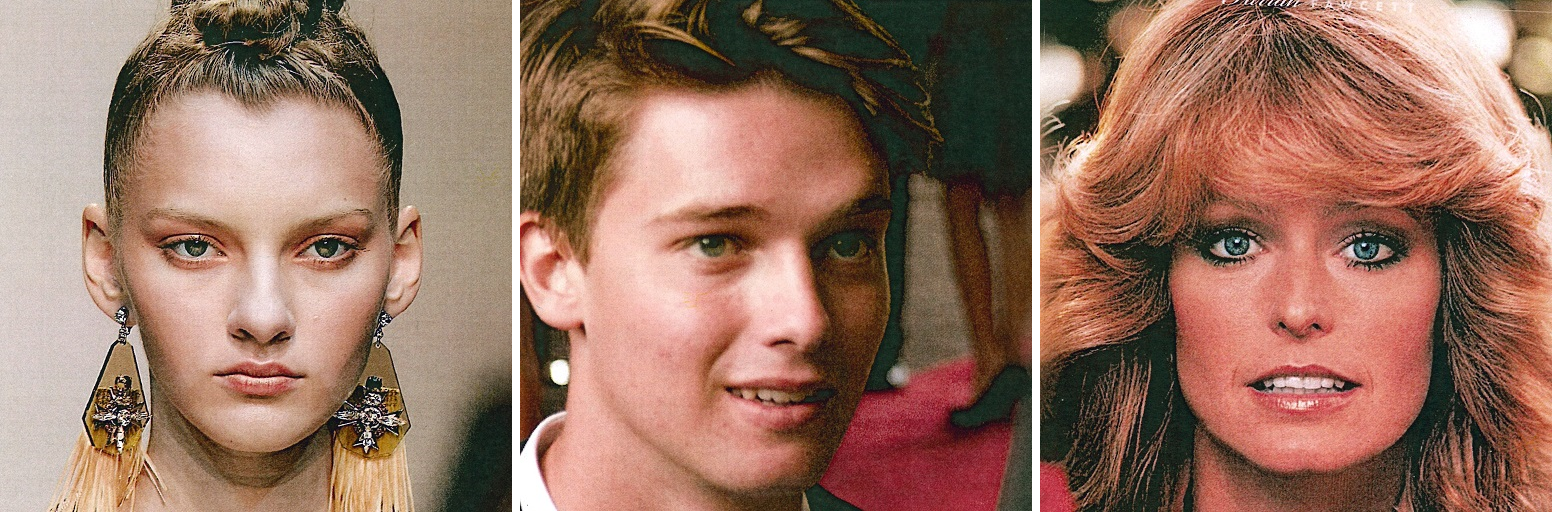}
    \label{fig:ProGAN-a}
\end{subfigure}
\hfill
\begin{subfigure}{.6\textwidth}
    \caption{PS samples produced by ProGAN}
    \includegraphics[width=\textwidth]{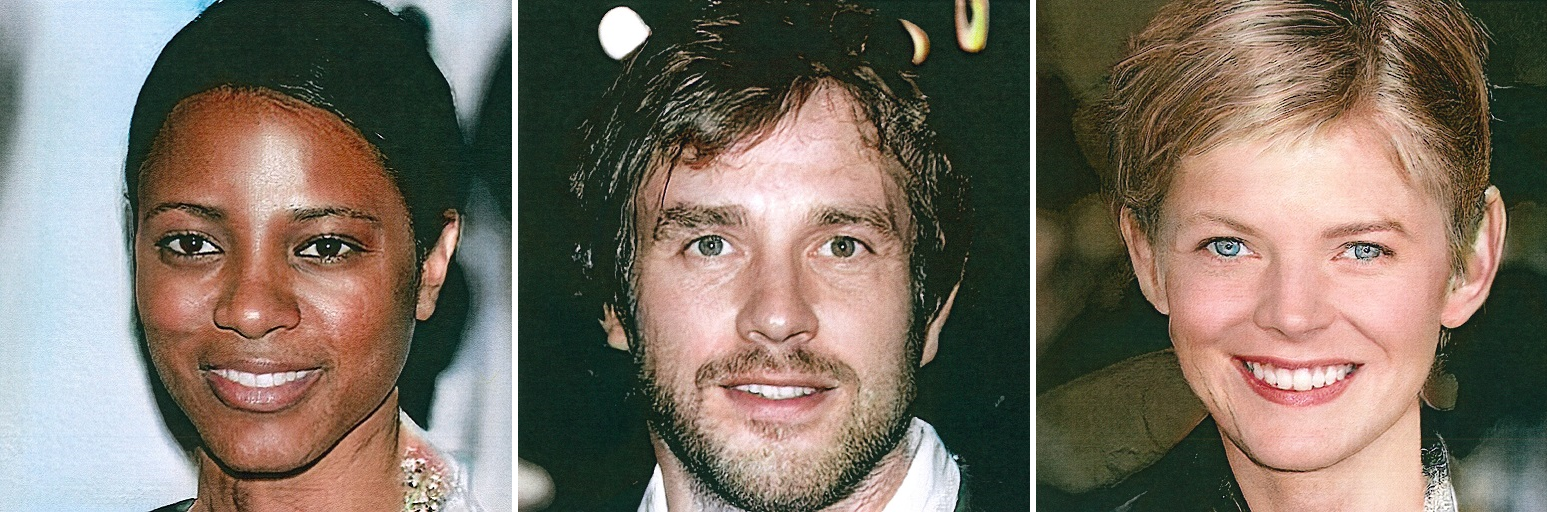}
    \label{fig:ProGAN-b}
\end{subfigure}
\label{fig:ProGAN}
\end{figure}



\begin{figure}[hbt!]
\centering
\caption{Some raw samples and StyleGAN generated samples from VIPPrint dataset}
\begin{subfigure}{.6\textwidth}
    \caption{Raw PS samples}
    \includegraphics[width=\textwidth]{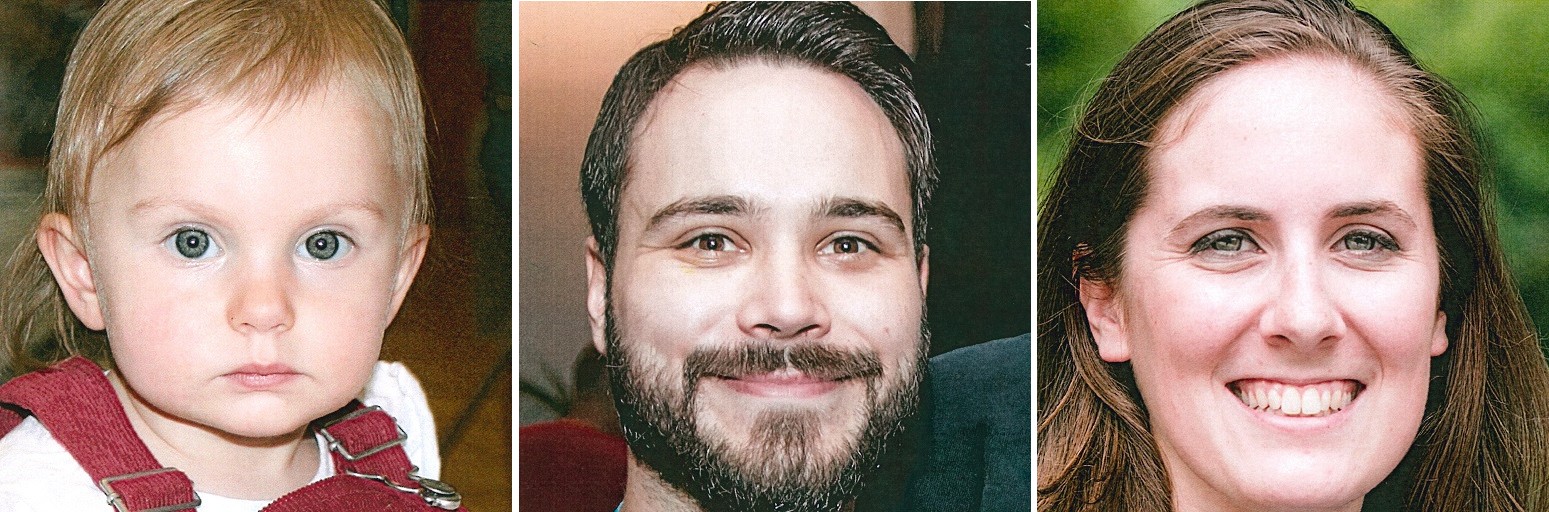}
    \label{fig:StyleGAN2-a}
\end{subfigure}
\hfill
\begin{subfigure}{.6\textwidth}
    \caption{PS samples produced by StyleGAN2}
    \includegraphics[width=\textwidth]{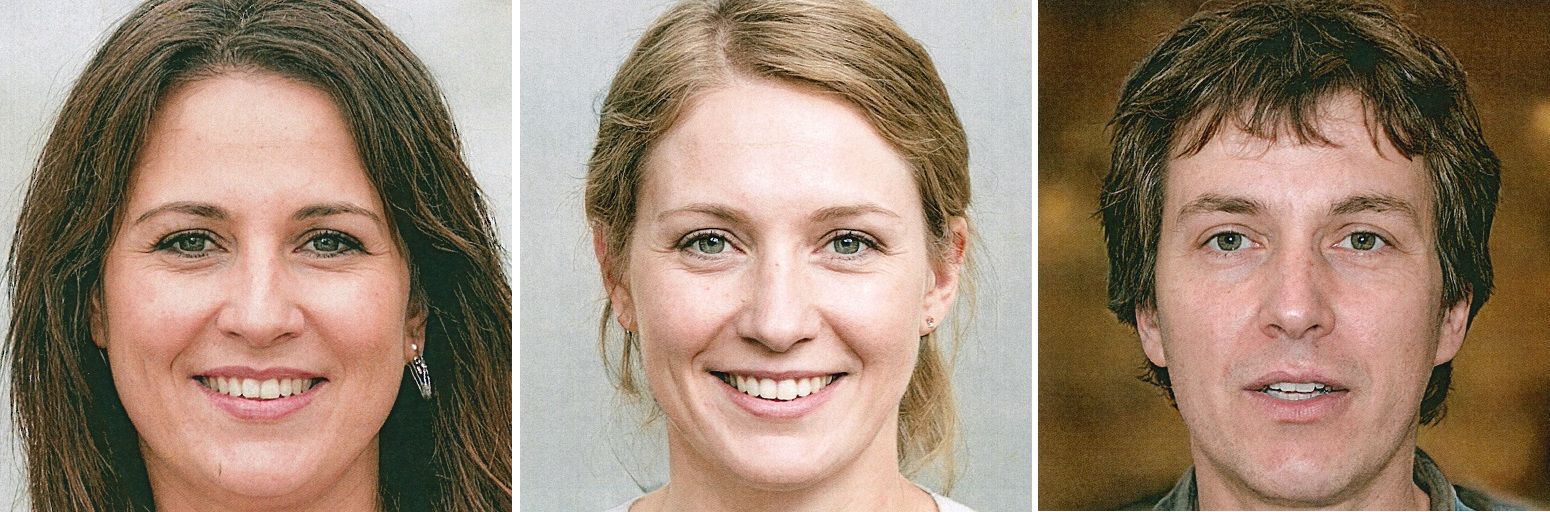}
    \label{fig:StyleGAN2-b}
\end{subfigure}
\label{fig:StyleGAN2}
\end{figure}


\begin{figure}[hbt!]
\centering
\caption{Raw and synthetic PS image and their zoomed regions}
\begin{subfigure}{.45\textwidth}
    \caption{Raw PS sample and its zoomed area}
    \includegraphics[width=\textwidth]{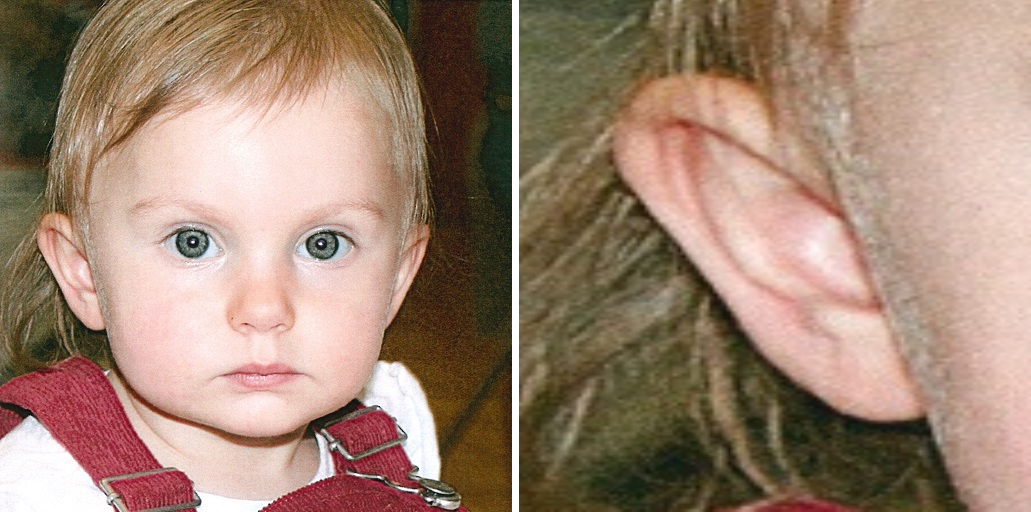}
    \label{fig:Zoomed_StyleGAN-a}
\end{subfigure}
\hfill
\begin{subfigure}{.45\textwidth} 
    \caption{StyleGAN2 generated PS sample and its zoomed area}
    \includegraphics[width=\textwidth]{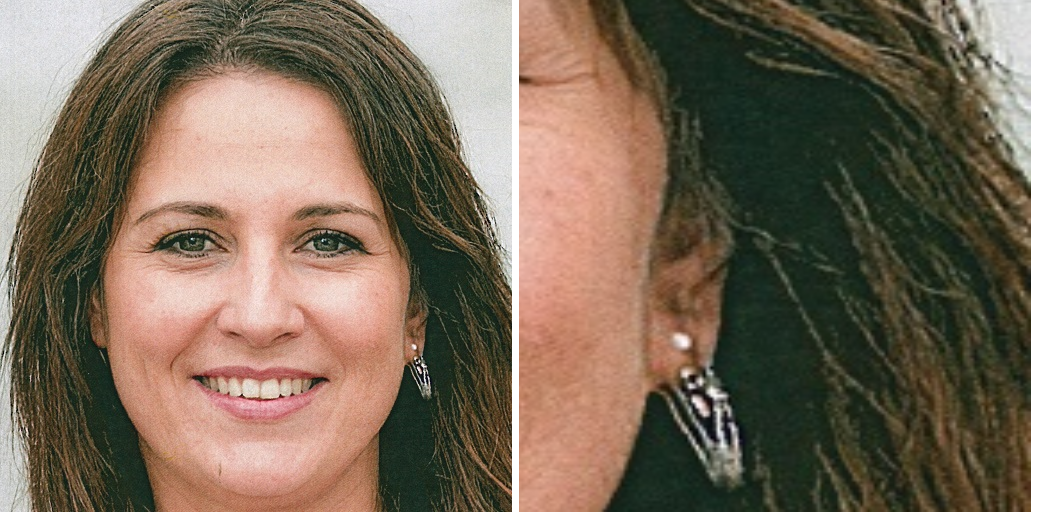}
    \label{fig:Zoomed_StyleGAN-b}
\end{subfigure}
\label{fig:Zoomed_StyleGAN}
\end{figure}

\subsection{Iris Extraction}
\label{sec:Iris Extraction}

As described, synthetic images generated from GANs have discrepancies in their irises which is not available in real images. this phenomenon is visible in Figure\mbox{~\ref{fig:raw_zoomed}} and Figure\mbox{~\ref{fig:StyleGAN2_zoomed}}. Figure\mbox{~\ref{fig:StyleGAN2_zoomed}} shows fake PS face images generated by StyleGAN2 with their zoomed left and right irises where Figure\mbox{~\ref{fig:raw_zoomed}} shows for the raw images. Irises that are zoomed from pristine images have a round shape, while synthetic ones generated from GANs have some distortions. Moreover, the light effects inside the iris between the left and right eye in synthetic images have mismatches, while the light reflection effects between the left and right pristine images have the same pattern. In this section, we describe how we extracted irises from real and fake PS face images. we provided an automated procedure to extract irises. After detecting irises from faces and extracting them, at the end of the Iris extraction step, we have our pristine and fake PS face images into Iris images.

\begin{figure}[hbt!]
    \centering
    \caption{some fake PS face samples generated by StyleGAN2 with their zoomed left and right irises
}
    \includegraphics[width=0.7\textwidth] {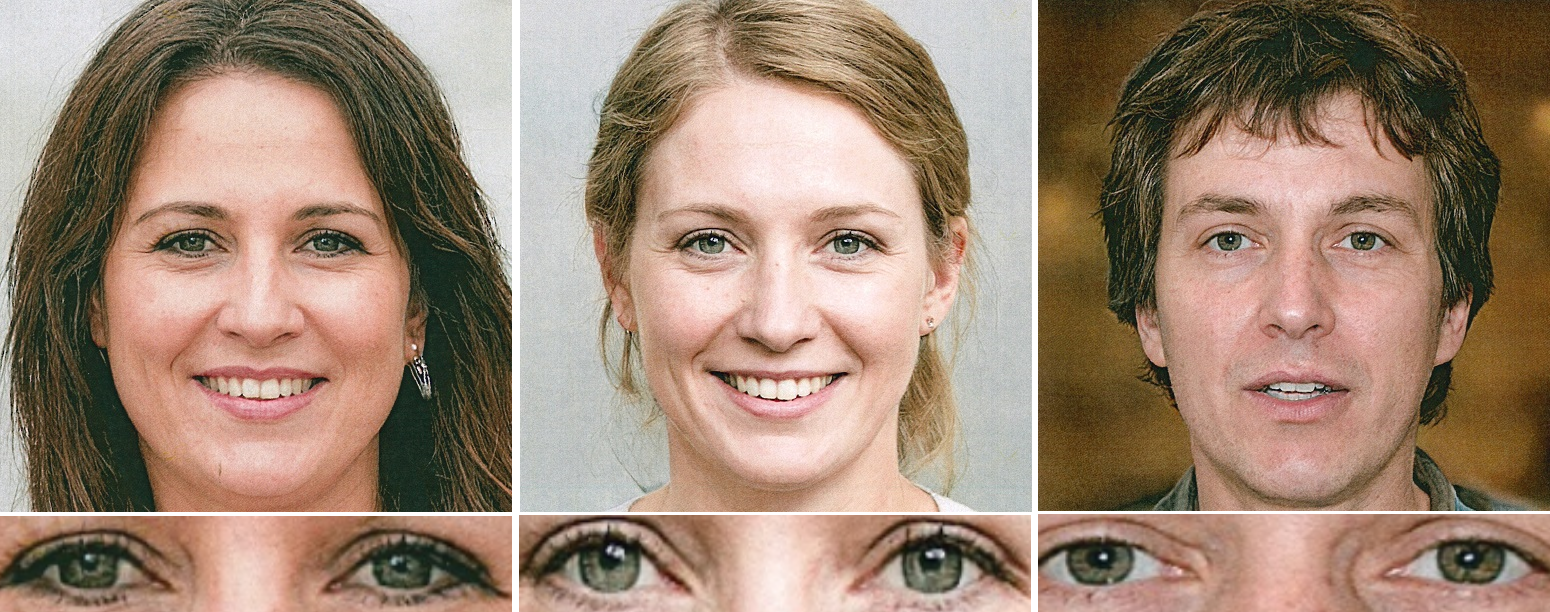}
    
    \captionsetup{justification=centering}
    \label{fig:raw_zoomed}
\end{figure}

\begin{figure}[hbt!]
    \centering
    \caption{Some raw samples with their zoomed left and right irises
}
    \includegraphics[width=0.7\textwidth]{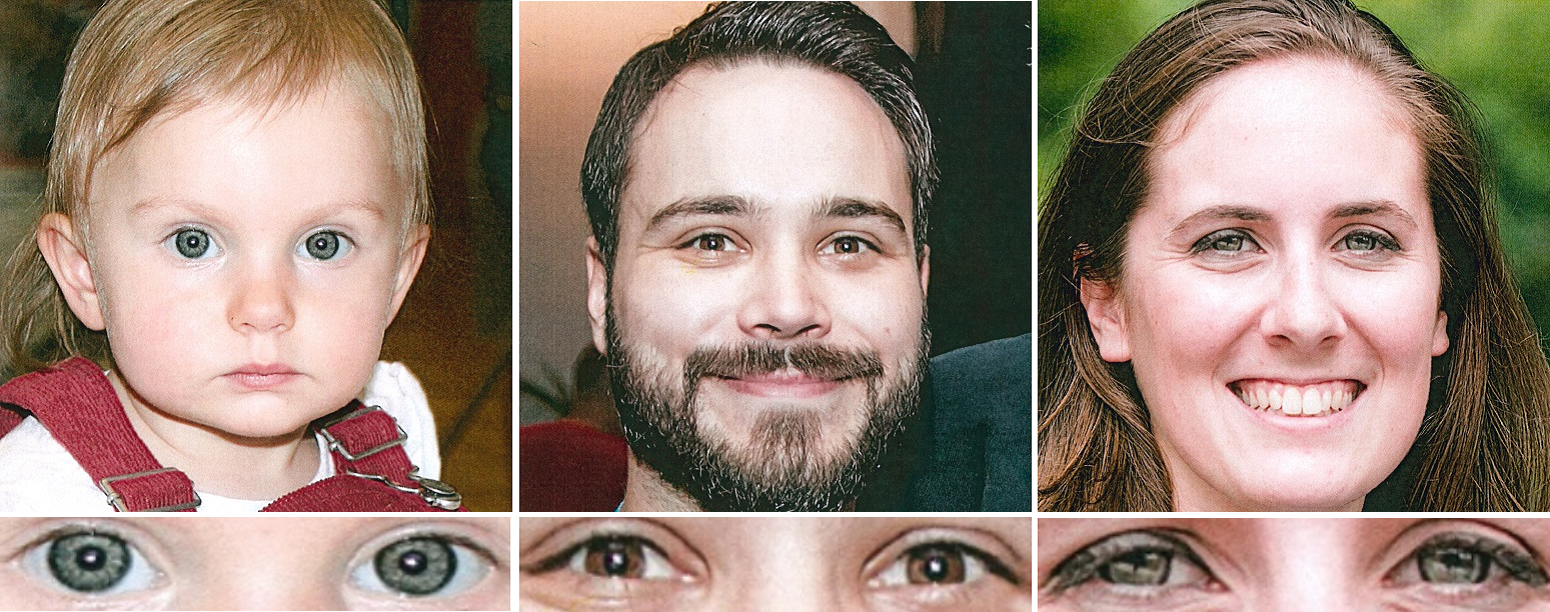}
    
    \captionsetup{justification=centering}
    \label{fig:StyleGAN2_zoomed}
\end{figure}

\begin{figure}[hbt!]
    \centering
    \caption{Image categorization after iris extraction step
}
    \includegraphics[width=1.0\textwidth]{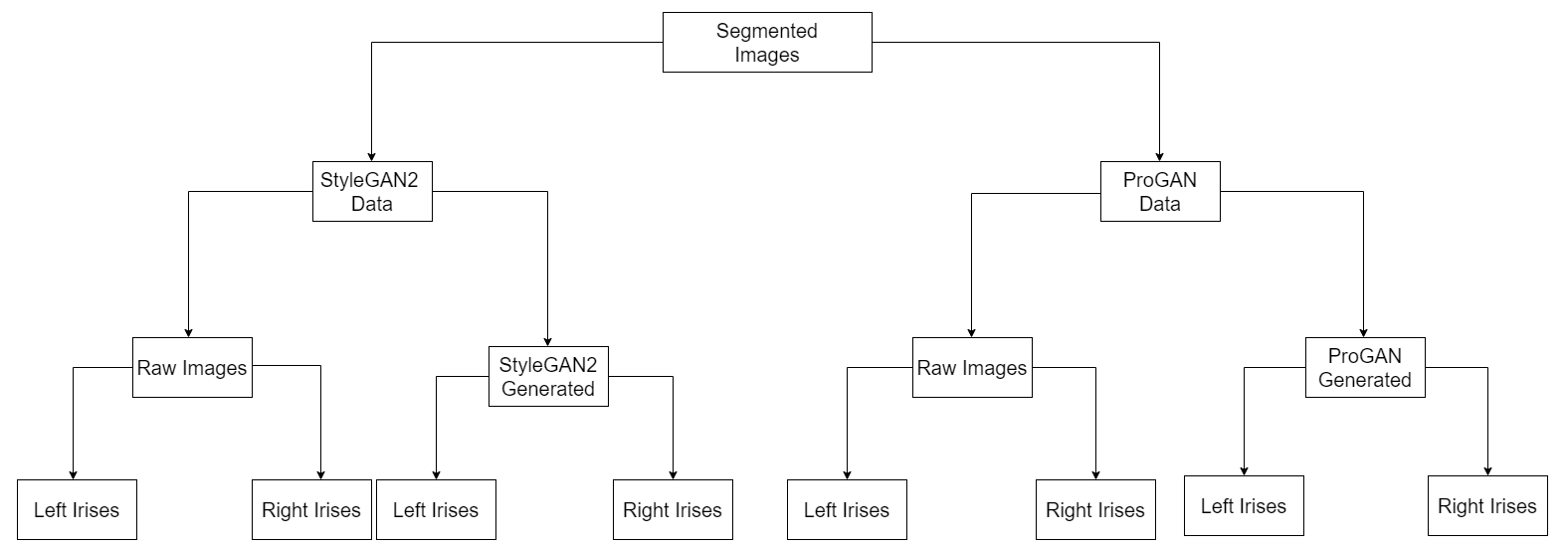}
    
    \captionsetup{justification=centering}
    \label{fig:cat}
\end{figure}

\subsection{Iris Reconstruction}
\label{sec:Iris Reconstruction}

Irises resulted from previous step, both pristine and synthetic iris images, may have some levels of irregularity. This irregularity can be due to the occlusion while capturing the real images or while generating where the model’s deficiency can add some more irregularities. this irregularity is present specifically when GAN models are applied{~\cite{NEURIPS2021_076ccd93}}.  As we know, Irises should be circular and to have intact iris shapes, we require an additional step. this additonal process is necessary since CNN models require similar circular left and right eye for performing the right analysis. This additional step tries to reconstruct the impaired irises applying image inpainting. One pristine and synthetic sample before and after reconstruction are shown respectively in Figure\mbox{~\ref{fig:real_reconstruction}} and\mbox{~\ref{fig:fake_reconstruction}}. In both figures, upper irises are resulted from after reconstruction, where bottom images represent before reconstruction which are not circular. Left images show the left irises, while the right images represent right irises and in both this phenomenon is visible. Reconstruction of iris images, especially when they are PS is a challenging task which has not been done in the literature so far. this study as far as we know, is the first study that applies a reconstruction phase as well for detecting pristine and synthetic iris images. 

\begin{figure*}[ht!]
    \centering
    \caption{Left and Right iris Samples Before and After Reconstruction.}
    \begin{subfigure}[t]{0.35\textwidth}
        \centering
        \caption{Pristine Irises}
        \includegraphics[width=\textwidth]{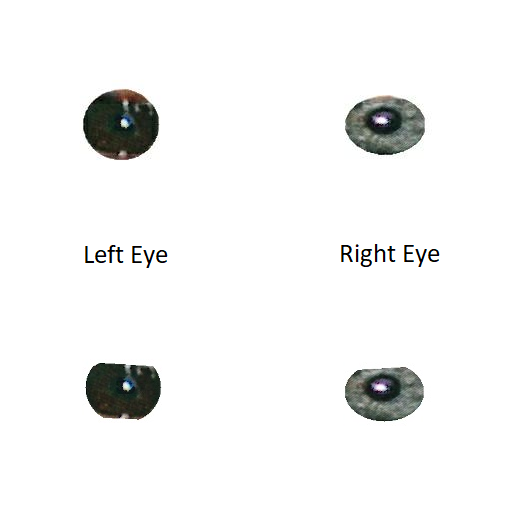}
        \label{fig:real_reconstruction}
    \end{subfigure}%
    ~ 
    \begin{subfigure}[t]{0.35\textwidth}
        \centering
        \caption{Synthetic Irises}
        \includegraphics[width=\textwidth]{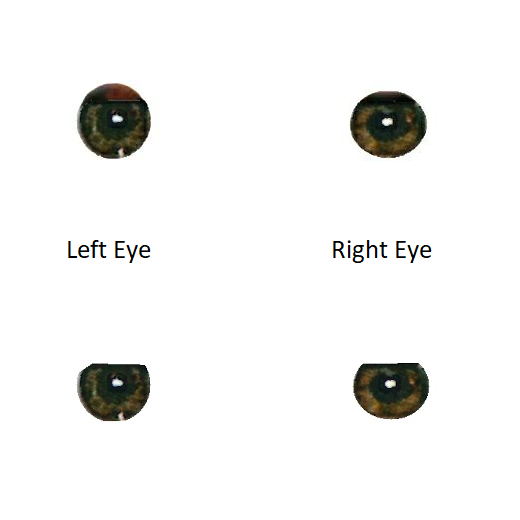}
        \label{fig:fake_reconstruction}
    \end{subfigure}
\end{figure*}

\subsection{Siamese Neural Networks}

In the literature, SNNs are frequently utilized for change detection. The goal of change detection is to determine if the target subject in the image changes over time. This modification ought to demonstrate how the subject's actual character and form had changed. In several areas, including security, privacy, urban planning, catastrophe prevention, and agricultural growth, change detection has been applied{~\cite{9884862}\cite{nowroozi2022spritz}\cite{nowroozi2022resisting}\cite{barni2020effectiveness}}.

 In the verification step, we consider SNN to verify whether the reconstructed irises are from pristine images or are the result of synthetic GAN-generated images \cite{nowroozi2023detecting}\cite{conti2021not}. we apply a Siamese network to calculate the distance between the left and right irises. Between left and right synthetic irises, there are higher levels of irregularities in comparison with pristine left and right irises. This irregularity is visible in the similarity score calculated from the Siamese model, which is illustrated in the following sections. Various pre-trained convolutional neural networks, including Resnet50, VGG16, MobileNet-v2, and Xception, have been used as the backbone of the SNN architecture. The effectiveness of each backbone in generating the similarity score to distinguish pristine and synthetic PS iris images has been examined as well when considering different backbones. 
A model architecture having two parallel neural networks that are comparable is known as an SNN, which is shown in Figure\mbox{~\ref{fig:Siamese_Network}}. The weights are shared between these two networks, and the networks have identical configurations, weights, and parameters. Each network receives a unique input, and their outputs are merged in order to provide some predictions. The core concept behind Siamese networks is that they are able to acquire the helpful data descriptors needed to contrast inputs from different subnetworks{~\cite{9116915}}. When there are several categories that are unknown during training or where there are few training examples for each category, the approach can be utilized for recognition or verification applications. The goal is to learn a function that converts input patterns into a target space such that the "semantic" distance in the input space is roughly approximated by the L1 norm in the target space. When the technique is used to perform a face verification task, during the learning phase it is trying to minimize a discriminative loss function that causes the similarity metric to be small for pairs of inputs from the same source and high for pairings from different sources. A convolutional network with an architecture built for robustness to geometric distortions maps the raw space to the target space{~\cite{1467314}}. As illustrated in the Figure\mbox{~\ref{fig:Siamese_Network}}, \(X_1\) and \(X_2\) are two different inputs, where both of them will be encoded by applying SNN through various layers. The encoding functions related to Inputs \(X_1\) and \(X_2\) are respectively \(f(X_1)\) and \(f(X_2)\). Whether two inputs are coming from the same source or not is based on the distance between the two functions \(f(X_1)\) and \(f(X_2)\). This distance is known as the "Euclidean" distance, which is considered also as an energy function and is calculated based on Formula{~\ref{eqn:Euclidean}}. Consequently, if \(X_1\) and \(X_2\) are similar, the value of E will be lower.

\begin{equation}
\label{eqn:Euclidean}
   E_w(X_1, X_2) = \|f_w(x_1) - f_w(x_2)\|
\end{equation}

\begin{figure}[!h]
  \captionsetup{justification=centering}
  \caption{Siamese Neural Network (SNN)}
  \includegraphics[width=0.8\linewidth]{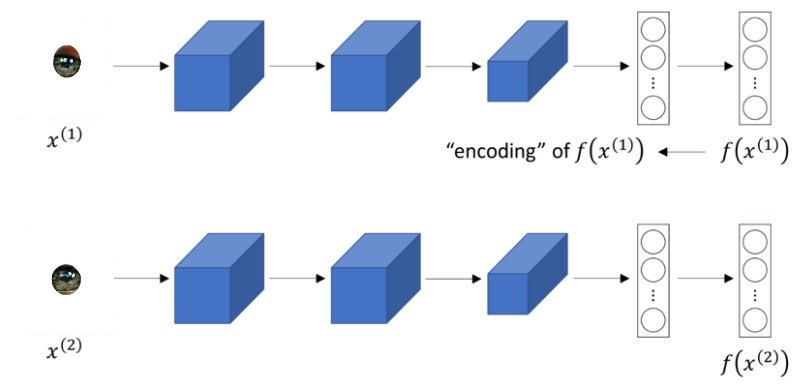}
  \label{fig:Siamese_Network}
\end{figure}

\section{ Experimental Results}
We applied SNN model to investigate the similarity of left and right irises resulted from previous steps. The results of similarity measure for raw and GAN-generated images are reported in the following, where Tables\mbox{~\ref{tab:result_raw_progan}}, \mbox{~\ref{tab:result_raw_stylegan}}, \mbox{~\ref{tab:result_gan_progan}}, and\mbox{~\ref{tab:result_gan_stylegan}} show respectively the similarity measure resulted from real images of ProGAN, real images of StyleGAN, GAN-generated images of ProGAN, and GAN-generated images of StyleGAN.

Tables\mbox{~\ref{tab:result_raw_progan}} and \mbox{~\ref{tab:result_raw_stylegan}} represent the similarity between left and right irises captured from real faces, Tables \mbox{~\ref{tab:result_gan_progan}} and\mbox{~\ref{tab:result_gan_stylegan}} show the values for GAN-generated irises. Comparing the results reveal that when irises are coming from real sources, the similarity measure is high and is in the range of 85.5560 and 94.9585, but when they are generated with GAN, the similarity range drops to the range of 56.5275 and 76.5844. These scores reveal the high level of distortion or irregularity present in irises when they are synthetic. This high level of asymmetry difference which is captured in our results can introduce a robust countermeasure for detecting GAN generated face images.
This study has been done by considering different backbones for our SNN model to further investigate the strength of each architecture for distinguishing pristine and fake PS iris images. MobileNet-v2 architecture as a backbone result in the highest value for similarity score for raw images, 95.04\% and 93.84\% respectively for raw images of ProGAN and StyleGAN. On the other hand, Xception produced lowest average similarity score for raw images of ProGAN, and second-lowest for raw images of StyleGAN. Comparison of similarity measure ranges between StyleGAN and ProGAN, respectively (56\% - 62\%) and (56\% - 76\%), reveals that StyleGAN is more powerful in synthesizing iris images and all SNN models resulted more similarity score, yet much lower than what models perceive as real iris. We also provided more information regarding different architectures which are presented in the tables. Among all, Xception model has the highest test accuracy for all four datasets. Computation complexity of Resnet50 is higher than other architectures since it has a greater number of parameters in comparison with other backbones. From time complexity we can order them from fastest to slowest respectively MobileNet-v2, Vgg16, Xception, and ResNet50.

\begin{table}[H]
\centering
\caption{Comparison of different Deep Learning models on RAW-ProGAN Images}
\label{tab:result_raw_progan}
\renewcommand{\arraystretch}{2.3}
\hspace*{-5mm}
\begin{adjustbox}{width=1\textwidth}
\captionsetup{justification=centering}
\begin{tabular}{|c|c|c|c|c|c|c|c|}
\hline
\textbf{CNN Models}   & \textbf{\begin{tabular}[c]{@{}c@{}}Training\\ Parameters\end{tabular}} & \textbf{\begin{tabular}[c]{@{}c@{}}Training\\ Loss\end{tabular}} & \textbf{\begin{tabular}[c]{@{}c@{}}Training\\ Accuracy\end{tabular}} & \textbf{\begin{tabular}[c]{@{}c@{}}Computation Time\\ (mins)\end{tabular}} & \textbf{\begin{tabular}[c]{@{}c@{}}Testing\\ Loss\end{tabular}} & \textbf{\begin{tabular}[c]{@{}c@{}}Testing\\ Accuracy\end{tabular}} & \textbf{\begin{tabular}[c]{@{}c@{}}Similarity\\ (Avrg)\end{tabular}} \\ \hline
\textbf{ResNet50}     & 87,877,632                                                             & 0.1273                                                           & 0.8684                                                               & 69.04                                                                      & 0.1201                                                          & 0.8799                                                              & 94.9585                                                              \\ \hline
\textbf{Vgg16}        & 16,976,384                                                              & 0.1656                                                           & 0.8450                                                               & 10.92                                                                      & 0.1794                                                          & 0.8206                                                              & 90.4041                                                              \\ \hline
\textbf{MobileNet-v2} & 42,142,208                                                             & 0.1224                                                           & 0.8693                                                               & 7.09                                                                       & 0.1452                                                          & 0.8548                                                              & 95.0447                                                              \\ \hline
\textbf{Xception}     & 67,308,032                                                             & 0.0889                                                           & 0.9003                                                               & 24.196                                                                     & 0.0762                                                          & 0.9238                                                              & 85.5560                                                              \\ \hline
\end{tabular}
\end{adjustbox}

\end{table}

\begin{table}[!h]
\centering
\caption{Comparison of different Deep Learning models on RAW-StyleGAN Images}
\label{tab:result_raw_stylegan}
\renewcommand{\arraystretch}{2.3}
\hspace*{-5mm}
\begin{adjustbox}{width=1\textwidth}
\captionsetup{justification=centering}
\begin{tabular}{|c|c|c|c|c|c|c|c|}
\hline
\textbf{CNN Models}   & \textbf{\begin{tabular}[c]{@{}c@{}}Training\\ Parameters\end{tabular}} & \textbf{\begin{tabular}[c]{@{}c@{}}Training\\ Loss\end{tabular}} & \textbf{\begin{tabular}[c]{@{}c@{}}Training\\ Accuracy\end{tabular}} & \textbf{\begin{tabular}[c]{@{}c@{}}Computation Time\\ (mins)\end{tabular}} & \textbf{\begin{tabular}[c]{@{}c@{}}Testing\\ Loss\end{tabular}} & \textbf{\begin{tabular}[c]{@{}c@{}}Testing\\ Accuracy\end{tabular}} & \textbf{\begin{tabular}[c]{@{}c@{}}Similarity\\ (Avrg)\end{tabular}} \\ \hline
\textbf{ResNet50}     & 87,877,632                                                             & 0.1131                                                           & 0.8844                                                               & 67.67                                                                      & 0.1358                                                          & 0.8642                                                              & 93.8289                                                              \\ \hline
\textbf{Vgg16}        & 16,976,384                                                              & 0.2652                                                           & 0.7169                                                               & 15.38                                                                      & 0.2181                                                          & 0.7819                                                              & 91.8952                                                              \\ \hline
\textbf{MobileNet-v2} & 42,142,208                                                             & 0.1815                                                           & 0.8058                                                               & 9.9126                                                                     & 0.1726                                                          & 0.8274                                                              & 93.8425                                                              \\ \hline
\textbf{Xception}     & 67,308,032                                                             & 0.1094                                                           & 0.8954                                                               & 25.68                                                                      & 0.0919                                                          & 0.9081                                                              & 92.7738                                                              \\ \hline
\end{tabular}
\end{adjustbox}
\end{table}

\begin{table}[!h]
\centering
\caption{Comparison of different Deep Learning models on GAN-ProGAN Images}
\label{tab:result_gan_progan}
\renewcommand{\arraystretch}{2.3}
\hspace*{-5mm}
\begin{adjustbox}{width=1\textwidth}
\captionsetup{justification=centering}
\begin{tabular}{|c|c|c|c|c|c|c|c|}
\hline
\textbf{CNN Models}   & \textbf{\begin{tabular}[c]{@{}c@{}}Training\\ Parameters\end{tabular}} & \textbf{\begin{tabular}[c]{@{}c@{}}Training\\ Loss\end{tabular}} & \textbf{\begin{tabular}[c]{@{}c@{}}Training\\ Accuracy\end{tabular}} & \textbf{\begin{tabular}[c]{@{}c@{}}Computation Time\\ (mins)\end{tabular}} & \textbf{\begin{tabular}[c]{@{}c@{}}Testing\\ Loss\end{tabular}} & \textbf{\begin{tabular}[c]{@{}c@{}}Testing\\ Accuracy\end{tabular}} & \textbf{\begin{tabular}[c]{@{}c@{}}Similarity\\ (Avrg)\end{tabular}} \\ \hline
\textbf{ResNet50}     & 87,877,632                                                             & 0.4363                                                           & 0.5164                                                               & 27.39                                                                      & 0.4240                                                          & 0.5760                                                              & 58.2235                                                              \\ \hline
\textbf{Vgg16}        & 16,976,384                                                             & 0.2335                                                           & 0.7322                                                               & 12.39                                                                      & 0.2160                                                          & 0.7840                                                              & 62.9761                                                              \\ \hline
\textbf{MobileNet-v2} & 42,142,208                                                             & 0.1483                                                           & 0.8507                                                               & 12.3478                                                                    & 0.1631                                                          & 0.8369                                                              & 60.5014                                                              \\ \hline
\textbf{Xception}     & 67,308,032                                                             & 0.1026                                                           & 0.8947                                                               & 18.153                                                                     & 0.1105                                                          & 0.8895                                                              & 56.5275                                                              \\ \hline
\end{tabular}
\end{adjustbox}
\end{table}

\begin{table}[H]
\centering
\caption{Comparison of different Deep Learning models on GAN-StyleGAN Images}
\label{tab:result_gan_stylegan}
\renewcommand{\arraystretch}{2.3}
\hspace*{-5mm}
\begin{adjustbox}{width=1\textwidth}
\captionsetup{justification=centering}
\begin{tabular}{|c|c|c|c|c|c|c|c|}
\hline
\textbf{CNN Models}   & \textbf{\begin{tabular}[c]{@{}c@{}}Training\\ Parameters\end{tabular}} & \textbf{\begin{tabular}[c]{@{}c@{}}Training\\ Loss\end{tabular}} & \textbf{\begin{tabular}[c]{@{}c@{}}Training\\ Accuracy\end{tabular}} & \textbf{\begin{tabular}[c]{@{}c@{}}Computation Time\\ (mins)\end{tabular}} & \textbf{\begin{tabular}[c]{@{}c@{}}Testing\\ Loss\end{tabular}} & \textbf{\begin{tabular}[c]{@{}c@{}}Testing\\ Accuracy\end{tabular}} & \textbf{\begin{tabular}[c]{@{}c@{}}Similarity\\ (Avrg)\end{tabular}} \\ \hline
\textbf{ResNet50}     & 87,877,632                                                             & 0.1059                                                           & 0.9012                                                               & 48.18                                                                      & 0.1006                                                          & 0.8994                                                              & 76.5844                                                              \\ \hline
\textbf{Vgg16}        & 16,976,384                                                             & 0.5000                                                           & 0.5340                                                               & 8.433                                                                      & 0.5800                                                          & 0.5110                                                              & 69.4118                                                              \\ \hline
\textbf{MobileNet-v2} & 42,142,208                                                             & 0.2787                                                           & 0.7470                                                               & 11.01                                                                      & 0.5000                                                          & 0.5000                                                              & 64.0681                                                              \\ \hline
\textbf{Xception}     & 67,308,032                                                             & 0.0782                                                           & 0.9266                                                               & 24.67                                                                      & 0.0678                                                          & 0.9322                                                              & 56.7691                                                              \\ \hline
\end{tabular}
\end{adjustbox}
\end{table}

\section{Conclusion}
We proposed a novel PS iris image dataset that images are originally captured from face images of VIPPrint Printed and Scanned face dataset{~\cite{VIPP}}. Extracting and applying irises are a challenging task specifically when (i) the images are PS and not extracted directly from digital images since printing and scanning add some noises to the documents, and (ii) irises can be occluded due to many reasons, so before feeding them to our neural network, other post-processing steps are required. 
In the literature, there are some researches done on capturing the similarity between the left and right eye, more exact eye holes, like{~\cite{Wang2022}}, while this study focused on extracting irises instead of eyes, or eye holes, since (i) iris comprises the most import information required, (ii) as other irrelevant parts such as the sclera, or white of the eye, and skins around the eyes are excluded, computations are lower and more efficient, (iii) and the similarity score better represent the difference captured as other parts are almost similar on different sides of the face.

The underlying assumption for this study comes from the fact that the synthetic irises generated with GAN models have some level of irregularity. In this study, we considered an SNN model with 4 different backbones including ResNet50, Xception, VGG16, and MobileNet-v2 measuring the similarity score of the left and right irises in order to distinguish the pristine irises from synthetic ones.
The similarity scores resulted from all four SNN architectures reveal the promising performance of our solution since there is a high gap between the similarity score of pristine PS iris images and GAN-generated images, whether from ProGAN or StyleGAN. The similarity scores resulted from StyleGAN are higher than ProGAN architecture, but at its highest it is 76\%, while for the pristine images are ranging from 85\% to 95\%. The best SNN model from the accuracy, both training and test,  point of view was Xception.


\section*{Acknowledgement}
We are thankful to Seyedsadra Shoari for his help in building our proposed dataset.

\bibliographystyle{splncs04}
\bibliography{Ref}

\end{document}